\def\year{2022}\relax
\documentclass[letterpaper]{article} 
\usepackage{aaai22}  
\usepackage{times}  
\usepackage{helvet}  
\usepackage{courier}  
\usepackage[hyphens]{url}  
\usepackage{graphicx} 
\usepackage{natbib}  
\usepackage{caption} 
\DeclareCaptionStyle{ruled}{labelfont=normalfont,labelsep=colon,strut=off} 
\usepackage{booktabs}       
\usepackage{xcolor}         
\usepackage{multirow}
\usepackage{caption}
\usepackage{amsfonts}
\usepackage{mathtools}
\usepackage{amsmath}
\usepackage[ruled,vlined]{algorithm2e}
\usepackage{subcaption}

\newcommand{\ignore}[1]{}
\newcommand{\para}[1]{\smallskip\noindent\textbf{#1}}

\title{Attributing Fair Decisions with Attention Interventions}
\author{
   Ninareh Mehrabi, Umang Gupta, Fred Morstatter, Greg Ver Steeg, Aram Galstyan
}
\affiliations{
Information Sciences Institute, University of Southern California \\
\texttt{\{ninarehm,umanggup,morstatt\}@usc.edu}, \texttt{\{gregv,galstyan\}@isi.edu}

}

\usepackage{bibentry}

\begin{document}

\maketitle

\begin{abstract}

The widespread use of Artificial Intelligence (AI) in consequential domains, such as healthcare and parole decision-making systems, has drawn intense scrutiny on the fairness of these methods. However, ensuring fairness is often insufficient as the rationale for a contentious decision needs to be audited, understood, and defended. We propose that the attention mechanism can be used to ensure fair outcomes while simultaneously providing feature attributions to account for how a decision was made. Toward this goal, we design an attention-based model that can be leveraged as an attribution framework. It can identify features responsible for both performance and fairness of the model through attention interventions and attention weight manipulation. Using this attribution framework, we then design a post-processing bias mitigation strategy and compare it with a suite of baselines. We demonstrate the versatility of our approach by conducting experiments on two distinct data types, tabular and textual.
\end{abstract}

\section{Introduction}


Machine learning algorithms that optimize for performance (e.g., accuracy) often result in unfair outcomes~\cite{10.1145/3457607}. These algorithms capture biases present in the training datasets causing discrimination toward different groups. As machine learning continues to be adopted into fields where discriminatory treatments can lead to legal penalties, fairness and interpretability have become a necessity and a legal incentive in addition to an ethical responsibility~\cite{barocas2016big,hacker2020explainable}. Existing methods for fair machine learning include applying complex transformations to the data so that resulting representations are fair~\cite{gupta2021controllable,NEURIPS2018_415185ea,roy2019mitigating,advforg, pmlr-v89-song19a}, adding  regularizers to incorporate fairness~\cite{pmlr-v54-zafar17a,kamishima2012fairness,mehrabi2020statistical}, or modifying the outcomes of unfair machine learning algorithms to ensure fairness~\cite{NIPS2016_9d268236}, among others. Here we present an alternative approach \footnote{Code can be found at: \url{https://github.com/Ninarehm/Attribution}}, which works by identifying the significance of different features in causing unfairness and reducing their effect on the outcomes using an attention-based mechanism.

With the advancement of transformer models and the attention mechanism \cite{NIPS2017_3f5ee243}, recent research in Natural Language Processing (NLP) has tried to analyze the effects and the interpretability of the attention weights on the decision making process~\cite{wiegreffe-pinter-2019-attention,jain-wallace-2019-attention,serrano-smith-2019-attention,hao2021self}. Taking inspiration from these works, we propose to use an attention-based mechanism to study the fairness of a model. The attention mechanism provides an intuitive way to capture the effect of each attribute on the outcomes. Thus, by introducing the attention mechanism, we can analyze the effect of specific input features on the model's fairness. We form visualizations that explain model outcomes and help us decide which attributes contribute to accuracy vs.\ fairness. We also show and confirm the observed effect of indirect discrimination in previous work \cite{zliobaite2015survey,6175897,ijcai2017-549} in which even with the absence of the sensitive attribute, we can still have an unfair model due to the existence of proxy attributes. Furthermore, we show that in certain scenarios those proxy attributes contribute more to the model unfairness than the sensitive attribute itself.

Based on the above observations, we propose a post-processing bias mitigation technique by diminishing the weights of features most responsible for causing unfairness. We perform studies on datasets with different modalities and show the flexibility of our framework on both tabular and large-scale text data, which is an advantage over existing interpretable non-neural and non-attention-based models. Furthermore, our approach provides a competitive and interpretable baseline compared to several recent fair learning techniques.

To summarize, the contributions of this work are as follows: (1) We propose a new framework for attention-based classification in tabular data, which is interpretable in the sense that it allows to quantify the effect of each attribute on the outcomes; (2) We then use these attributions to study the effect of different input features on the fairness and accuracy of the models; (3) Using this attribution framework, we propose a post-processing bias mitigation technique that can reduce unfairness and provide competitive accuracy vs. fairness trade-offs; (4) Lastly, we show the versatility of our framework by applying it to large-scale non-tabular data such as text.

\section{Approach}

In this section, we describe our classification model that incorporates the attention mechanism. It can be applied to both text and tabular data and is inspired by works in attention-based models in text-classification~\cite{zhou-etal-2016-attention}. We incorporate attention over the input features. Next, we describe how this attention over features can attribute the model's unfairness to certain features. Finally, using this attribution framework, we propose a post-processing approach for mitigating unfairness.

In this work, we focus on binary classification tasks. We assume access to a dataset of triplets $\mathcal D = \{x_i , y_i , a_i \}^N_{i=1}$, where $x_i,y_i,a_i$ are i.i.d. samples from data distribution $p(\mathbf x, \mathbf y, \mathbf a)$. $\mathbf a \in \{a_1,\ldots a_l\}$ is a discrete variable with $l$ possible values and denotes the sensitive or protected attributes with respect to which we want to be fair, $\mathbf y\in \{0,1\}$ is the true label, $\mathbf x\in \mathbb{R}^m $ are features of the sample which may include sensitive attributes. We use $\hat y_o$ to denote the binary outcome of the original model, and $\hat y_z^k$ will represent the binary outcome of a model in which the attention weights corresponding to $k^\text{th}$ feature are zeroed out. Our framework is flexible and general that it can be used to find attribution for any fairness notion. More particularly, we work with the group fairness measures like \textit{Statistical Parity}~\cite{Dwork:2012:FTA:2090236.2090255}, \textit{Equalized Odds}~\cite{NIPS2016_9d268236}, and \textit{Equality of Opportunity}~\cite{NIPS2016_9d268236}, which are defined as:\footnote{We describe and use the  definition of these fairness measures as implemented in \textit {Fairlearn }package~\cite{bird2020fairlearn}. } 

\noindent
\textbf{Statistical Parity Difference (SPD)}:
$$
\text{SPD}(\hat {\mathbf y}, \mathbf a ) =
    \max_{a_i, a_j}|P(\hat{\mathbf y}=1\mid \mathbf a=a_i)-P(\hat{\mathbf y}=1\mid \mathbf a =a_j)|
$$
\textbf{Equality of Opportunity Difference (EqOpp)}:
\begin{align*}
\text{EqOpp}(\hat {\mathbf y}, \mathbf a, \mathbf y) =
    \max_{a_i, a_j} |P(\hat{\mathbf y}=1\mid\mathbf a = a_i,\mathbf y=1)&\\-P(\hat{\mathbf y}=1\mid\mathbf a = a_j,\mathbf y=1)|
\end{align*}
\textbf{Equalized Odds Difference (EqOdd)}:
\begin{align*}
\text{EqOdd}(\hat {\mathbf y}, \mathbf a, \mathbf y) = 
\max_{a_i, a_j} \max_{y\in\{0,1\}} |P(\hat{\mathbf y}=1\mid\mathbf a = a_i,\mathbf y=y)&\\- P(\hat{\mathbf y}=1\mid\mathbf a = a_j,\mathbf y=y)|
\end{align*}

\begin{figure}[h]
    \centering
    \begin{subfigure}[b]{0.21\textwidth}
        \includegraphics[width=0.96\textwidth,trim=0cm 0cm 0cm 0cm,clip=true]{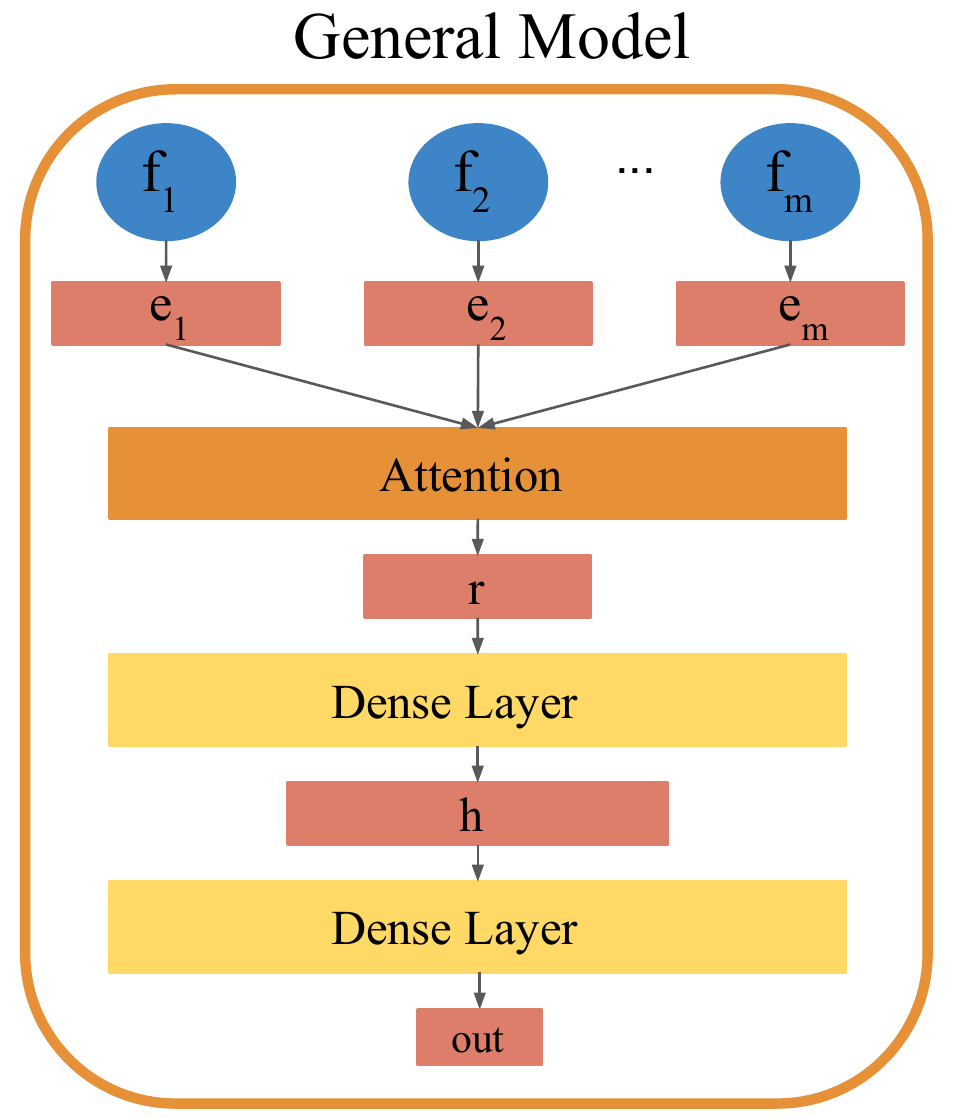}
        \caption{Classification model.}
        \label{fig:original_model}
    \end{subfigure}
    \begin{subfigure}[b]{0.24\textwidth}
        \includegraphics[width=\textwidth,trim=0cm 0cm 0cm 0cm,clip=true]{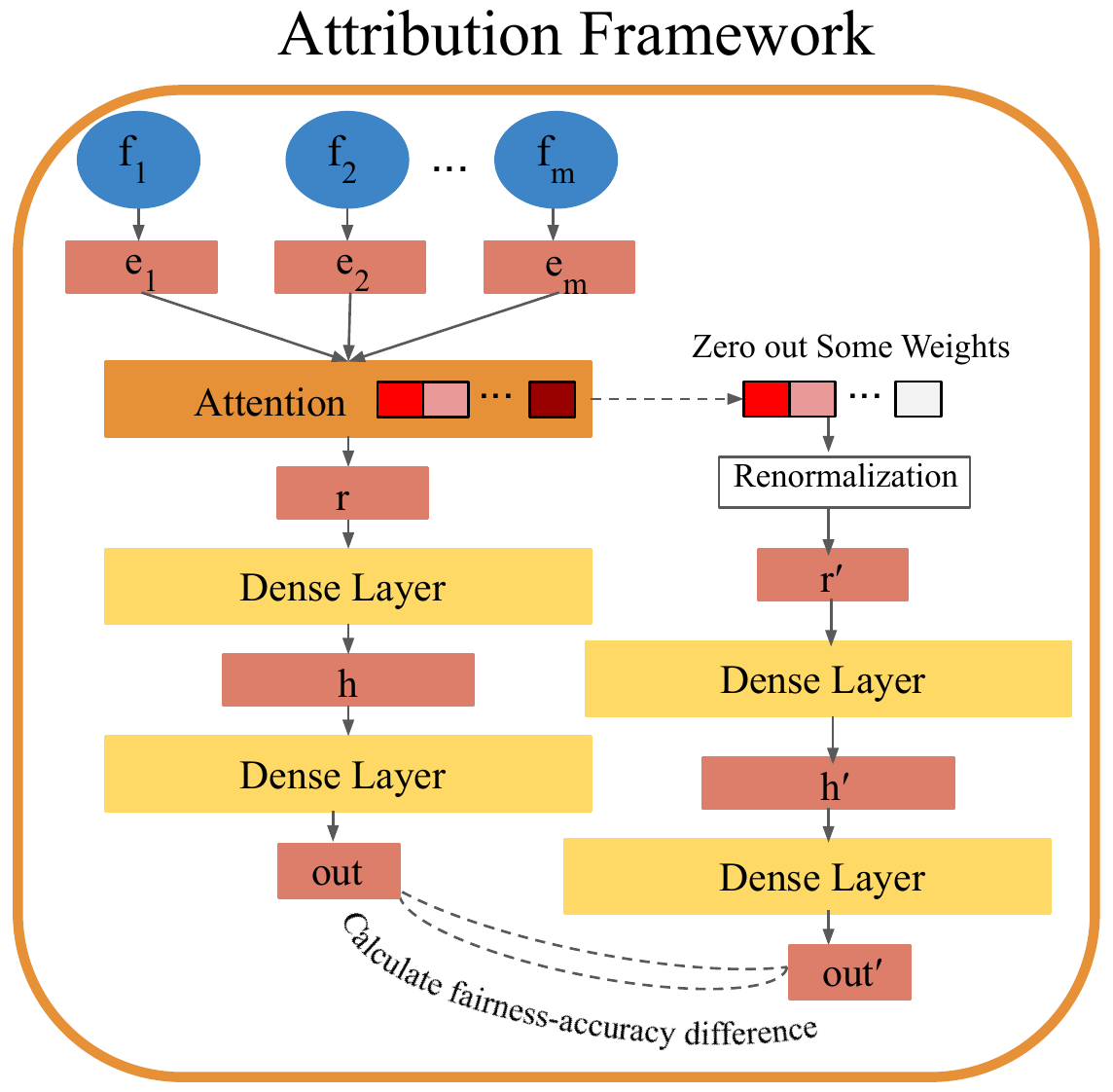}
        \caption{Attribution framework.}
        \label{fig:intervention_model}
    \end{subfigure}
    \caption{(a) In general classification model, for each feature $f_k$ a vector representation $e_k$ of length $d^e$ is learned. This is passed to the attention layer which produces a $d^e$-dimensional vector representation for the sample instance $i$ which is passed to two dense layers to get the final classification output. (b) The Attribution framework has the same architecture as the general model. One outcome is obtained through the original model and another through the model that has some attention weights zeroed. The observed difference in accuracy and fairness measures will indicate the effect of the zeroed out features on accuracy and fairness.}
    \label{fig:model}
\end{figure}


\begin{algorithm}[h]
\SetAlgoLined
Input: decay rate $d_r$ ($0 \leq d_r <1$), $n$ test samples indexed by variable $i$. \\
Output: final predictions, unfair features. \\
Calculate the attention weights $\alpha_{ki}$ for $k^{\text{th}}$ feature in sample $i$ using the attention layer as in Eq.~\ref{eq:attention_op1}.\\
{unfair\_feature\_set} = \{\} \\
\For{each feature (index) $k$}{
    \If{$\text{SPD}(\hat {\mathbf y}_o, \mathbf {a})- \text{SPD}(\hat {\mathbf y}_z^k, \mathbf {a}) \geq 0$}{
    {unfair\_feature\_set}  = {unfair\_feature\_set}  $\cup\   \{k\}$
    }
}
\For{each feature (index) $k$}{
\If{$k$ in  \text{unfair\_feature\_set}}{
Set $\alpha_{ki} \leftarrow (d_r \times \alpha_{ki})$ for all $n$ samples
}
}
Use new attention weights 
to obtain the final predictions $\hat Y$. \\
\Return $\hat Y$,  {unfair\_feature\_set}
 \caption{Bias Mitigation with Attention}
 \label{mitig_alg}
\end{algorithm}

\subsection{General Model: Incorporating Attention over Inputs in Classifiers}

We consider each feature value as an individual entity (like the words are considered in text-classification) and learn a fixed-size embedding $\{e_k\}_{k=1}^m,\ e_k \in \mathbb{R}^{d^e}$ for each feature, $\{f_k\}_{k=1}^m$. These vectors are passed to the attention layer. The Computation of  attention weights and the final representation for a sample is described in Eq.~\ref{eq:attention_op1}.  $E = [e_1 \ldots e_m], \ E \in \mathbb R ^{d^e\times m}  $ is the concatenation of all the embeddings, $w \in  \mathbb{R}^{d^e} $ is a learnable parameter, $r  \in \mathbb{R}^{d^e}$ denotes the overall sample representation, and  $\alpha \in \mathbb{R}^{m}$ denotes the attention weights.
%
%
\begin{align}
H        = \tanh(E)  \quad \label{eq:attention_op1}
\alpha   = \text{softmax}(w^TH) \quad
r        = \tanh(E\alpha^T)
\end{align}
The resulting representation, $r$, is passed to the feed-forward layers for classification. In this work, we have used 
two feed-forward layers (See Fig.~\ref{fig:model} for overall architecture).
\subsection{Fairness Attribution with Attention Weights}
The aforementioned classification model with the attention mechanism combines input feature embeddings by taking a weighted combination. By manipulating the weights, we can intuitively capture the effects of specific features on the output.  To this end, we observe the effect of each attribute on the fairness of outcomes by zeroing out or reducing its attention weights and recording the change. Other works have used similar ideas to understand the effect of attention weights on accuracy and evaluate interpretability of the attention weights by comparing the difference in outcomes in terms of measures such as Jensen-Shannon Divergence~\cite{serrano-smith-2019-attention} but not for fairness. We are interested in the effect of features on fairness measures.  Thus, we measure the difference in fairness of the outcomes based on the desired fairness measure. A large change in fairness measure and a small change in performance of the model would indicate that this feature is mostly responsible for unfairness, and it can be dropped without causing large impacts on performance. The overall framework is shown in Fig.~\ref{fig:model}. First, the outcomes are recorded with the original attention weights intact (Fig.~\ref{fig:original_model}). Next, attention weights corresponding to a particular feature are zeroed out, and the difference in performance and fairness measures is recorded (Fig.~\ref{fig:intervention_model}). Based on the observed differences, one may conclude how incorporating this feature contributes to fairness/unfairness.

To measure the effect of the $k^{th}$ feature on different fairness measures, we consider the difference in the fairness of outcomes of the original model and model with $k^{th}$ feature's effect removed. For example, for statistical parity difference, we will consider $\text{SPD}(\hat {\mathbf y}_o, \mathbf {a})- \text{SPD}(\hat {\mathbf y}_z^k, \mathbf {a})$. A negative value will indicate that the $k^{th}$ feature helps mitigate unfairness, and a positive value will indicate that the $k^{th}$ feature contributes to unfairness. This is because $\hat{y}_z^k$ captures the exclusion of the $k^{th}$ feature (zeroed out attention weight for that feature) from the decision-making process. If the value is positive, it indicates that not having this feature makes the bias lower than when we include it. Notice here, we focus on global attribution, so we measure this over all the samples; however, this can also be turned into local attribution by focusing on individual sample $i$ only.
\subsection{Mitigating Bias by Removing Unfair Features} \label{mitig_sec}
As discussed in the previous section, we can identify features that contribute to unfair outcomes according to different fairness measures. A simple technique to mitigate or reduce bias is to reduce the attention weights of these features. This mitigation technique is outlined in Algorithm~\ref{mitig_alg}. In this algorithm, we first individually set attention weights for each of the features in all the samples to zero and monitor the effect on the desired fairness measure. We have demonstrated the algorithm for SPD, but other measures, such as EqOdd, EqOpp, and even accuracy can be used (in which case the ``unfair\_feature\_set'' can be re-named to feature set which harms accuracy instead of fairness). If the $k^{th}$ feature contributes to unfairness, we reduce its attention weight using decay rate value. This is because $\hat {\mathbf y}_z^k$ captures the exclusion of the $k^{th}$ feature (zeroed attention weight for that feature) compared to the original outcome $\hat {\mathbf y}_o$ for when all the feature weights are intact; otherwise, we use the original attention weight. We can also control the fairness-accuracy trade-off by putting more attention weight on features that boost accuracy while keeping the fairness of the model the same and down-weighting features that hurt accuracy, fairness, or both. 

This post-processing technique has a couple of advantages over previous works in bias mitigation or fair classification approaches. First, the post-processing approach is computationally efficient as it does not require model retraining to ensure fairness for each sensitive attribute separately. Instead, the model is trained once by incorporating all the attributes, and then one manipulates attention weights during test time according to particular  needs and use-cases. Second, the proposed mitigation method provides an explanation and can control the fairness-accuracy trade-off. 
This is because manipulating the attention weights reveals which features are important for getting the desired outcome, and by how much.
This provides an explanation for the outcome and also a mechanism to control the fairness-accuracy trade-off by the amount of the manipulation.

\section{Experimental Setup}
We perform a suite of experiments on synthetic and real-world datasets to evaluate our attention based interpretable fairness framework. The experiments on synthetic data are intended to elucidate interpretability in controlled settings, where we can manipulate the relations between input and output feature. The experiments on real-world data aim to validate the effectiveness of the proposed approach on both tabular and non-tabular (textual) data. Below we describe the experiments, datasets, and respective baselines.  

\subsection{Types of Experiments}
We enumerate the experiments and their goals as follows: \\
\noindent{\bf Experiment 1: Attributing Fairness with Attention} The purpose of this experiment is to demonstrate that our attribution framework can capture correct attributions of features to fairness outcomes. We present our results for tabular data in Sec.~\ref{sec:interpret_fairness}.\\
\noindent {\bf Experiment 2: Bias Mitigation via Attention Weight Manipulation}  In  this experiment, we seek to validate the proposed post-processing bias mitigation framework and compare it with various recent mitigation approaches. The results for real-world tabular data  are presented in Sec.~\ref{sec:bias_mitigation}.\\
\noindent{\bf Experiment 3: Validation on Textual Data} The goal of this experiment is to demonstrate the flexibility of the proposed attention-based method by conducting experiments on non-tabular, textual data. The results are presented in Sec.~\ref{sec:textual_data}.

\subsection{Datasets}
\newcommand{\bern}{\text{Ber}}

\subsubsection{Synthetic Data} \label{syn_data_sec}
To validate the attribution framework, we created two synthetic datasets in which we control how features interact with each other and contribute to the accuracy and fairness of the outcome variable.
These datasets capture some of the common scenarios, namely the data imbalance (skewness) and indirect discrimination issues, arising in fair decision or classification problems.

\paragraph{Scenario 1:} First, we create a simple scenario to demonstrate that our framework identifies correct feature attributions for fairness and accuracy. We create a feature that is correlated with the outcome (responsible for accuracy), a discrete feature that causes the prediction outcomes to be biased (responsible for fairness), and a continuous feature that is independent of the label or the task (irrelevant for the task). For intuition, suppose the attention-based attribution framework works correctly. In this case, we expect to see a reduction in accuracy upon removing (i.e., making the attention weight zero) the feature responsible  for the accuracy, reduction in bias upon removing the feature responsible for bias, and very little or no change upon removing the irrelevant feature. With this objective, we generated a synthetic dataset with three features, i.e., $x = [f_1, f_2, f_3]$ as follows\footnote{We use $x\sim \bern(p)$ to denote that $x$  is a Bernoulli random variable with $P(x=1)=p$.}:
\begin{align*}
    f_1 ~\sim \bern(0.9)
    \quad f_2\sim \bern(0.5) 
    \quad f_3\sim &\mathcal N (0,1) \\
    \quad y\sim \begin{cases}
        \bern(0.9) & \text{if } f_2=1\\
        \bern(0.1) & \text{if } f_2=0
    \end{cases}
\end{align*}
Clearly, $f_2$ has the most predictive information for the task and is responsible for accuracy. Here, we consider $f_1$ as the sensitive attribute. $f_1$ is an imbalanced feature that can bias the outcome and is generated such that there is no intentional correlation between $f_1$ and the outcome, $y$ or $f_2$. $f_3$ is sampled from a normal distribution independent of the outcome $y$, or the other features, making it irrelevant for the task.
Thus, an ideal classifier would be fair if it captures the correct outcome without being affected by the imbalance in $f_1$. However, due to limited data and skew in $f_1$,  there will be some undesired bias --- few errors when $f_1=0$ can lead to large statistical parity.

\paragraph{Scenario 2:} Using features that are not identified as sensitive attributes can result in unfair decisions due to their implicit relations or correlations with the sensitive attributes. This phenomenon is called indirect discrimination~\cite{zliobaite2015survey,6175897,ijcai2017-549}. We designed this synthetic dataset to demonstrate and characterize the behavior of our framework under indirect discrimination. Similar to the previous scenario, we consider three features. Here,  $f_1$ is considered as the sensitive attribute, and $f_2$ is correlated with $f_1$ and the outcome, $y$. The generative process is as follows:
\begin{align*}
    f_1\sim \begin{cases}
        \bern(0.9) & \text{if } f_2=1\\
        \bern(0.1) & \text{if } f_2=0
    \end{cases} 
    \quad f_2\sim\bern(0.5) &
    \quad f_3\sim \mathcal N (0,1) \\ 
    \quad y\sim \begin{cases}
        \bern(0.7) & \text{if } f_2=1\\ 
        \bern(0.3) & \text{if } f_2=0
    \end{cases} 
\end{align*}

In this case $f_1$ and $y$ are correlated with 
$f_2$. The model should mostly rely on $f_2$ for its decisions. 
However, due to the correlation between $f_1$ and $f_2$, we expect $f_2$ to affect both the accuracy and fairness of the model. Thus, in this case, indirect discrimination is possible. Using such a synthetic dataset, we demonstrate a) indirect discrimination and b) the need  to have  an attribution framework to reason about unfairness and not blindly focus on the sensitive attributes for bias mitigation.


\subsubsection{Real-world Datasets}
We demonstrate our approach on the following real-world datasets:\\
\noindent{\bf Tabular Datasets:} We conduct our experiments on two real-world tabular datasets often used to benchmark fair classification techniques --- \textit{UCI Adult}~\cite{Dua:2019} and \textit{Heritage Health}\footnote{https://www.kaggle.com/c/hhp} datasets. The \textit{UCI Adult} dataset contains census information about individuals, with the prediction task being whether the income of the individual is higher than \$50k or not. The sensitive attribute, in this case, is gender (male/female).  The \textit{Heritage Health} dataset contains patient information, and the task is to predict the Charleson Index (comorbidity index, which is a patient survival indicator). Each patient is grouped into one of the 9 possible age groups, and we consider this as the sensitive attribute.  We used the same pre-processing and train-test splits as in~\citet{gupta2021controllable}.\\
\noindent{\bf Non-Tabular or Text Dataset:}
To demonstrate the flexibility of our approach, we also experiment with a non-tabular, text dataset. We used the \textit{biosbias} dataset~\cite{de2019bias}. The dataset contains short bios of individuals. The task is to predict the occupation of the individual from their bio.
We utilized the bios from the year 2018 from the \texttt{2018\_34} archive and  considered two occupations for our experiments, namely, nurse and dentist. The dataset was split into 70-15-15 train, validation, and test splits. \citet{de2019bias} has demonstrated the existence of gender bias in this prediction task and showed that certain gender words are associated with certain job types (e.g., \textit{she} to nurse and \textit{he} to dentist). 

\subsection{Bias Mitigation Baselines} 
\label{sec:baselines}
We compared our bias mitigation approach to a number of recent state of the art methods. For our experiments with tabular data, we focus on methods that are specifically optimized to achieve statistical parity. Results for other fairness notions can be found in the appendix. For our baselines, we consider methods that learn representations of data so that information about sensitive attributes is eliminated.
\textit{\textbf{CVIB}}~\cite{NEURIPS2018_415185ea} realizes this objective through a conditional variational autoencoder, whereas \textit{\textbf{MIFR}}~\cite{pmlr-v89-song19a} uses a combination of information bottleneck term and adversarial learning to optimize the fairness objective. 
\textit{\textbf{FCRL}}~\cite{gupta2021controllable} optimizes information theoretic objectives that can be used to achieve good trade-offs between fairness and accuracy by using specialized contrastive information estimators. 
In addition to information-theoretic approaches, we also considered baselines that use adversarial learning such as \textit{\textbf{MaxEnt-ARL}}~\cite{roy2019mitigating}, \textit{\textbf{LAFTR}}~\cite{pmlr-v80-madras18a}, and \textit{\textbf{Adversarial Forgetting}}~\cite{advforg}. Note that in contrast to our approach, the baselines described above are not interpretable as they  are incapable of directly attributing features to fairness outcomes. For the textual data, we compare our approach with the debiasing technique proposed in~\citet{de2019bias}, which works by masking the gender-related words and then training the model on this masked data. 


\section{Results}
\subsection{Attributing Fairness with Attention} 
\label{sec:interpret_fairness}

First, we test our method's ability to capture correct attributions in controlled experiments with synthetic data (described in Sec.~\ref{syn_data_sec}). We also conduct a similar experiment with \textit{UCI Adult} and \textit{Heritage Health} datasets which can be found in the appendix. Fig.~\ref{fig:interpret_results} summarizes our results by visualizing the attributions, which we now discuss. 

In \textit{Scenario 1}, as expected, $f_2$ is correctly attributed to being responsible for the accuracy and removing it hurts the accuracy drastically. Similarly, $f_1$ is correctly shown to be responsible for unfairness and removing it creates a fairer outcome. Ideally, the model should not be using any information about $f_1$ as it is independent of the task, but it does. Therefore, by removing $f_1$, we can ensure that information is not used and hence outcomes are fair. Lastly, as expected, $f_3$ was the irrelevant feature, and its effects on accuracy and fairness are negligible. Another interesting observation is that $f_2$ is helping the model achieve fairness since its exclusion means that the model should rely on $f_1$ for decision making, resulting in more bias; thus, removing $f_2$ harms accuracy and fairness as expected.

In \textit{Scenario 2}, our framework captures the effect of indirect discrimination. We can see that removing $f_2$  reduces bias as well as accuracy drastically. This is because $f_2$ is the predictive feature, but due to its correlation with $f_1$, it can also indirectly affect the model's fairness. More interestingly, although $f_1$ is the sensitive feature, removing it does not play a drastic role in fairness or the accuracy. This is an important finding as it shows why removing $f_1$ on its own can not give us a fairer model due to the existence of correlations to other features and indirect discrimination. Overall, our results are intuitive and thus validate our assumption that attention-based framework  can provide reliable feature attributions for the fairness and accuracy of the model. 
\subsection{Attention as a Mitigation Technique}
\label{sec:bias_mitigation}
\begin{figure}[h]
\centering
\begin{subfigure}[b]{0.35\textwidth}
\includegraphics[width=\textwidth,trim=2.5cm 0cm 4.5cm 1.0cm,clip=true]{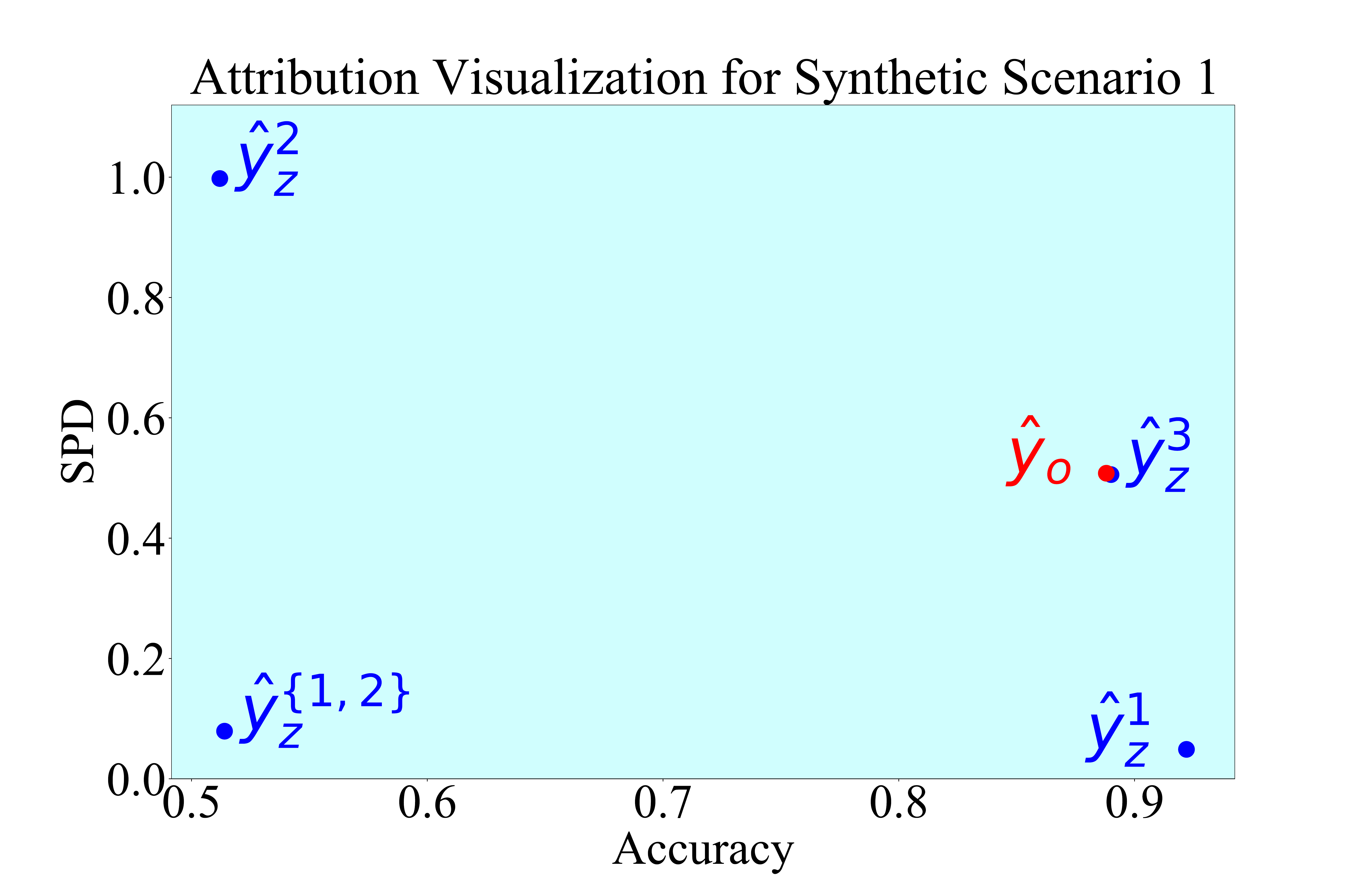}
\end{subfigure}
\begin{subfigure}[b]{0.35\textwidth}
\includegraphics[width=\textwidth,trim=2.5cm 0cm 4.5cm 1.0cm,clip=true]{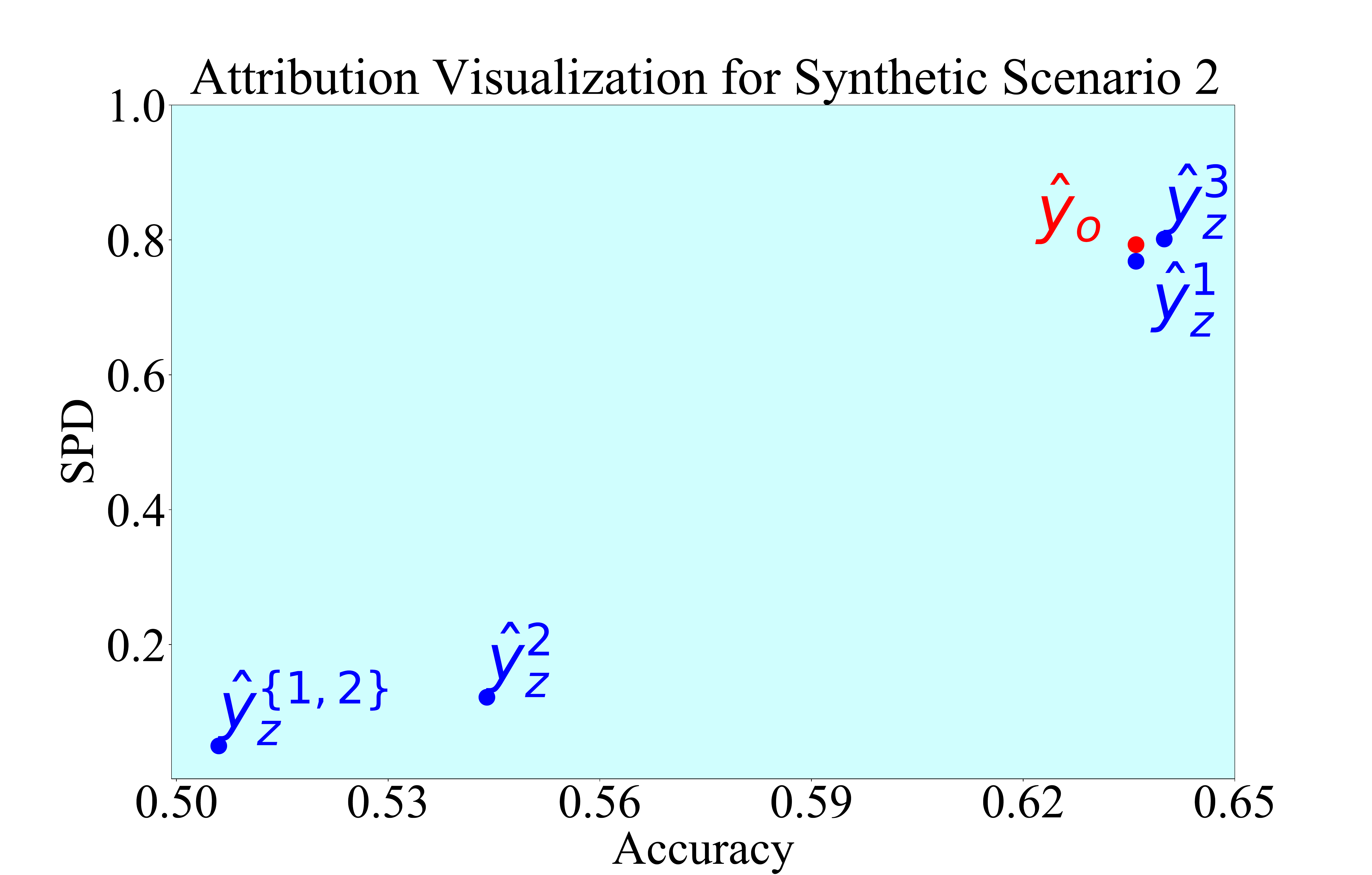}
\end{subfigure}
\caption{Results from the synthetic datasets. Following the $\hat{y}_o$ and $\hat{y}_z$ notations, $\hat{y}_o$ represents the original model outcome with all the attention weights intact, while $\hat{y}_z^k$ represents the outcome of the model in which the attention weights corresponding to $k^{th}$ feature are zeroed out (e.g. $\hat{y}_z^1$ represents when attention weights of feature $f_1$ are zeroed out). The results show how the accuracy and fairness of the model (in terms of statistical parity difference) change by exclusion of each feature.}
\label{fig:interpret_results}
\end{figure}
As we have highlighted earlier, understanding how the information within features interact and contribute to the decision making can be used to design effective bias mitigation strategies. One such example was shown in Sec.~\ref{sec:interpret_fairness}. Often real-world datasets have features which cause indirect discrimination, due to which fairness can not be achieved by simply eliminating the sensitive feature from the decision process. 
Using the attributions derived from our attention-based attribution framework, we propose a post-processing mitigation strategy. Our strategy is to intervene on  attention weights as discussed in Sec.~\ref{mitig_sec}. We first attribute and identify the features responsible for the unfairness of the outcomes, i.e., all the features whose exclusion will decrease the bias compared to the original model's outcomes and gradually decrease their attention weights to zero as also outlined in Algorithm~\ref{mitig_alg}. We do this by first using the whole fraction of the attention weights learned and gradually use less fraction of the weights until the weights are completely zeroed out. 


For all the  baselines described in Sec.~\ref{sec:baselines}, we used the approach outlined in~\citet{gupta2021controllable} for training a downstream classifier and evaluating the accuracy/fairness trade-offs. The downstream classifier was a 1-hidden-layer MLP with 50 neurons along with ReLU activation function. Our experiments were performed on Nvidia GeForce RTX 2080. Each method was trained with five different seeds, and we report the average accuracy and fairness measure as statistical parity difference (SPD). Results for other fairness notions can be found in the appendix. \textit{\textbf{CVIB}}, \textit{\textbf{MaxEnt-ARL}}, \textit{\textbf{Adversarial Forgetting}} and \textit{\textbf{FCRL}} are designed for statistical parity notion of fairness and are not applicable for other measures like Equalized Odds and Equality of Opportunity.  \textit{\textbf{LAFTR}} can only deal with binary sensitive attributes and thus not applicable for Heritage Health dataset. Notice that our approach does not have these limitations. For our approach, we vary the attention weights and report the resulting fairness-accuracy trade offs. 




\begin{figure*}[t]
\centering
    \begin{subfigure}[b]{0.43\textwidth}
        \includegraphics[width=\textwidth,trim=0cm 0cm 1.8cm 1.0cm,clip=true]{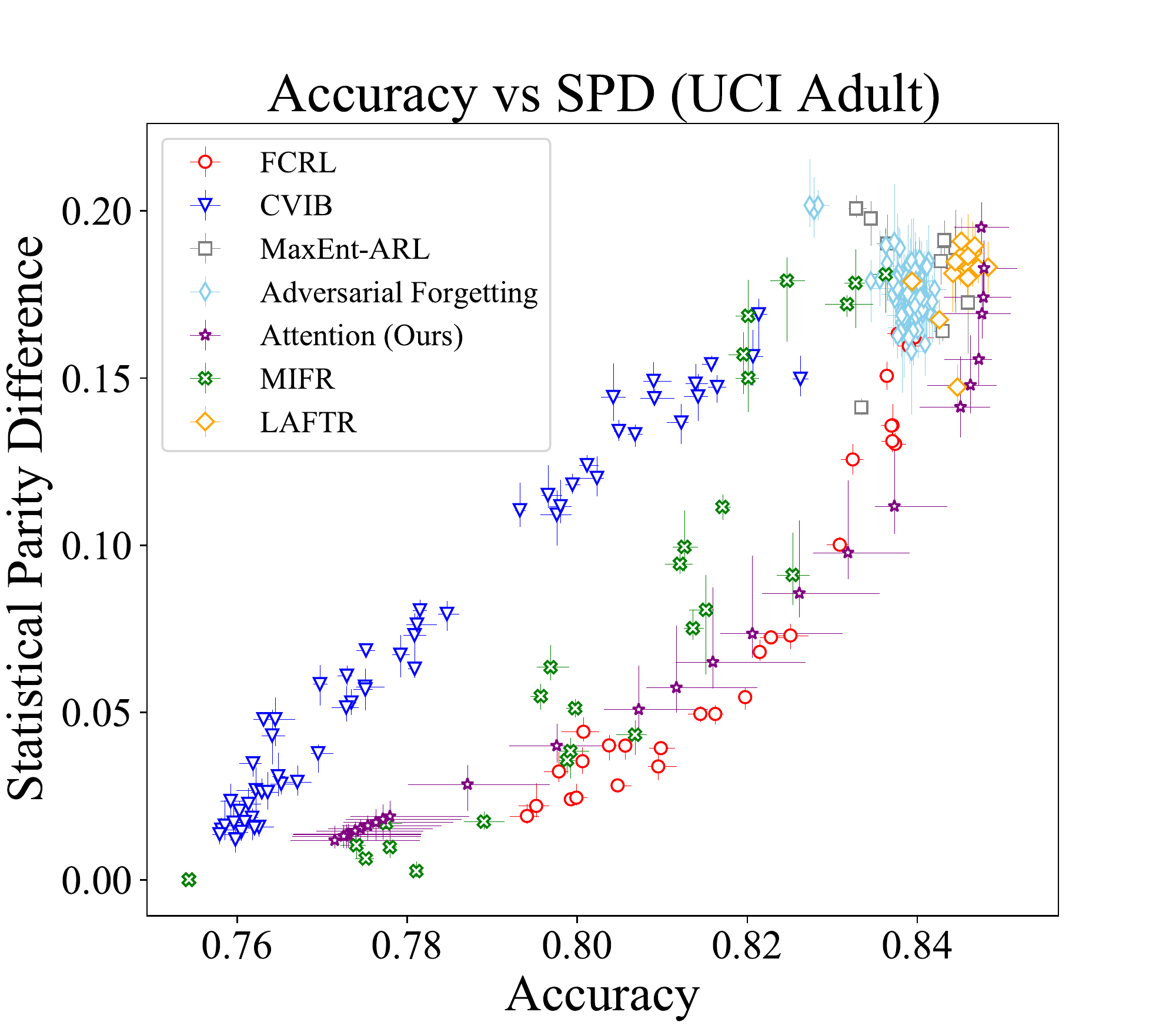}
    \end{subfigure}
    \begin{subfigure}[b]{0.43\textwidth}
        \includegraphics[width=\textwidth,trim=0cm 0cm 1.8cm 1cm,clip=true]{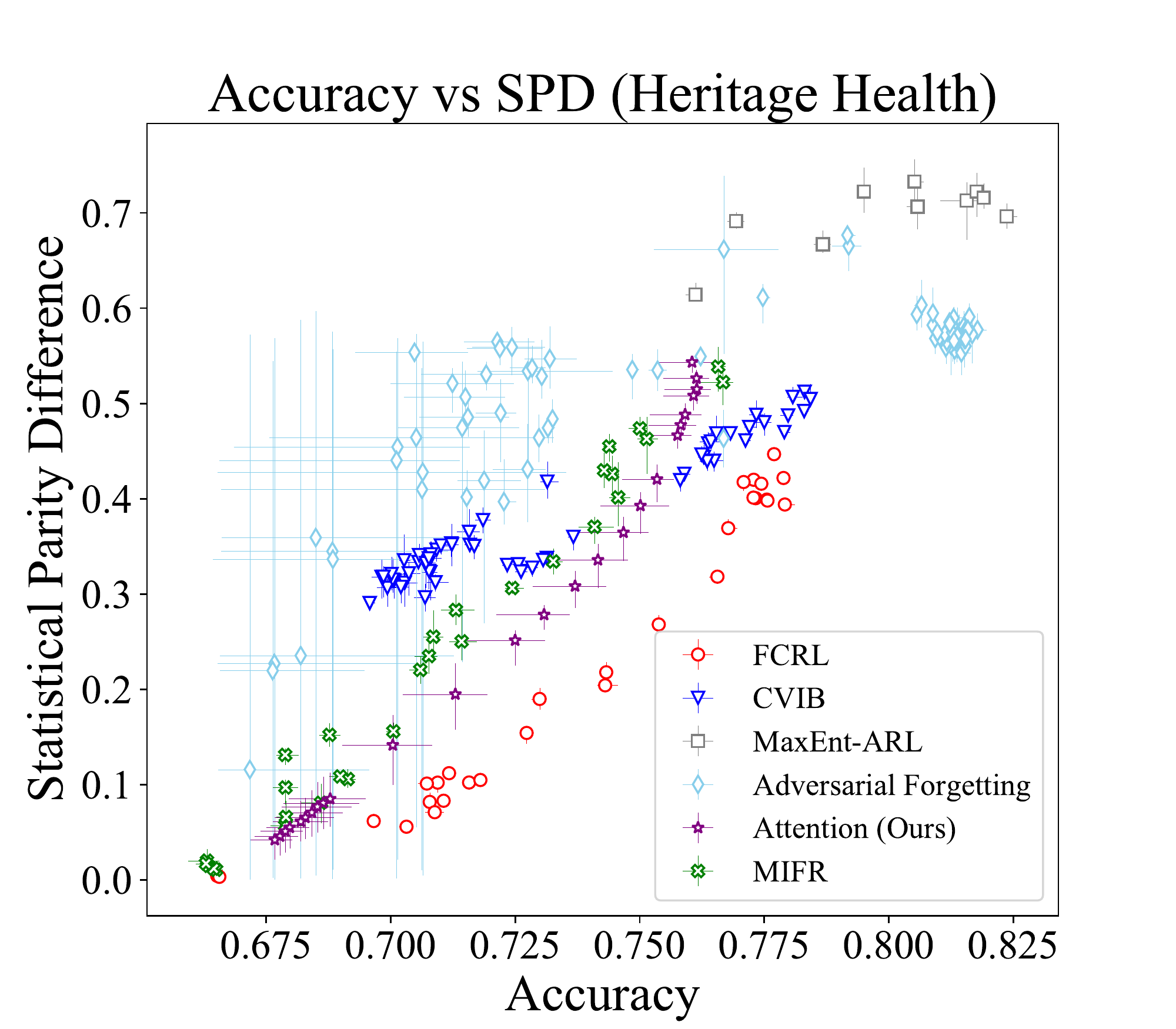}
    \end{subfigure}
    \caption{Accuracy vs parity curves for UCI Adult and Heritage Health datasets.}
    \label{fig:mitig_results}
\end{figure*}

Fig.~\ref{fig:mitig_results} compares fairness-accuracy trade-offs of different bias mitigation approaches. We desire outcomes to be fairer, i.e., lower values of \texttt{SPD} and to be more accurate, i.e., towards the right. The results show that using attention attributions can indeed be beneficial for reducing bias. Moreover, our mitigation framework based on the manipulation of the attention weights is competitive with state-of-the-art mitigation strategies. However, most of these approaches are  specifically designed and optimized to achieve parity and do not provide any interpretability. 
Our model can not only achieve comparable and competitive results, but it is also able to provide explanation such that the users exactly know what feature and by how much it was manipulated to get the corresponding outcome. Another advantage of our model is that it needs only one round of training. The adjustments to attention weights are made post-training; thus, it is possible to achieve different trade-offs. Moreover, our approach does not need to know sensitive attributes while training; thus, it could work with other sensitive attributes not known beforehand or during training.  Lastly, here we merely focused on mitigating bias (as our goal was to show that the attribution framework can identify problematic features and their removal would result in bias mitigation) and did not focus too much on improving accuracy and achieving the best trade-off curve which can be considered as the current limitation of our work. We manipulated attention weights of all the features that contributed to unfairness irrespective of if they helped maintaining high accuracy or not. However, the trade-off results can be improved by carefully considering the trade-off each feature contributes to with regards to both accuracy and fairness (e.g., using results from Fig.~\ref{fig:interpret_results}) to achieve better trade-off results which can be investigated as a future direction (e.g., removing problematic features that contribute to unfairness only if their contribution to accuracy is below a certain threshold value). The advantage of our work is that this trade-off curve can be controlled by controlling how many features and by how much to be manipulated which is not the case for most existing work.

\subsection{Experiments with Non-Tabular Data}
\label{sec:textual_data}

In addition to providing interpretability, our approach is flexible and useful for controlling fairness in modalities other than tabular datasets. To put this to the test, we applied our model to mitigate bias in text-based data. We consider the \textit{biosbias} dataset~\cite{de2019bias}, and  use our mitigation technique to reduce observed biases in the classification task performed on this dataset. We compare our approach with the debiasing technique proposed in the original paper~\cite{de2019bias}, which works  by masking the gender-related words and then training the model on this masked data. As discussed earlier, such a method is computationally inefficient. It requires  re-training the model or creating a new masked dataset, each time it is  required to debias the model against different attributes, such as gender vs. race. 
For the baseline pre-processing method, we masked the gender-related words, such as names and gender words, as provided in the \textit{biosbias} dataset and trained the model on the filtered dataset. On the other hand, we trained the model on the raw bios for our post-processing method and only manipulated attention weights of the gender words during the testing process as also provided in the \textit{biosbias} dataset. 
\begin{table*}[!t]
\centering
\begin{tabular}{ p{3.5cm} c c c}
 \toprule
\textbf{Method}&\textbf{Dentist TPRD (stdev)}&\textbf{Nurse TPRD (stdev)}&\textbf{Accuracy (stdev)}\\
 \midrule
 \parbox[t]{2mm}{\multirow{1}{*}{\shortstack[l]{Post-Processing (Ours)}}}
&\textbf{0.0202 (0.010)}&\textbf{0.0251 (0.020)}&0.951 (0.013)\\[0.5pt]

 \parbox[t]{2mm}{\multirow{1}{*}{\shortstack[l]{Pre-Processing}}}
&0.0380 (0.016)&0.0616 (0.025)&0.946 (0.011)\\[0.5pt]
\midrule
 \parbox[t]{2mm}{\multirow{1}{*}{\shortstack[l]{Not Debiased Model}}}
&0.0474 (0.025)&0.1905 (0.059)&\textbf{0.958 (0.011)}\\[0.5pt]


 \bottomrule
\end{tabular}
 \caption{Difference of the True Positive Rates (TPRD) amongst different genders for the dentist and nurse occupations on the biosbias dataset. Our introduced post-processing method is the most effective in reducing the disparity for both occupations compared to the pre-processing technique.}
\label{bios_results}
\end{table*}
\begin{figure*}[h]
    \centering
    \includegraphics[width=0.95\textwidth,trim=0cm 8cm 0cm 3.49cm,clip=true]{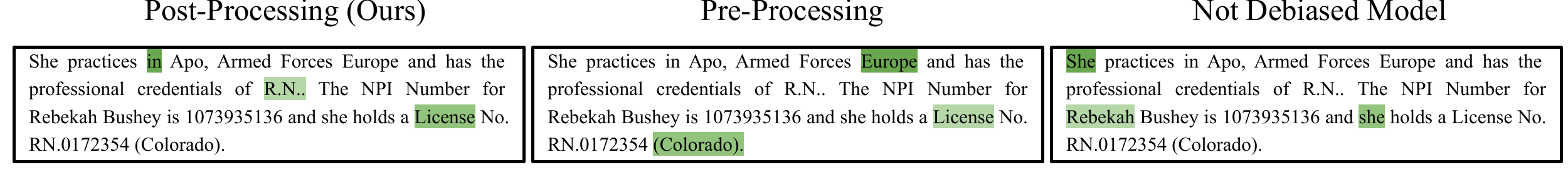}
     \includegraphics[width=0.95\textwidth,trim=0.24cm 7.8cm 1.4cm 3.49cm,clip=true]{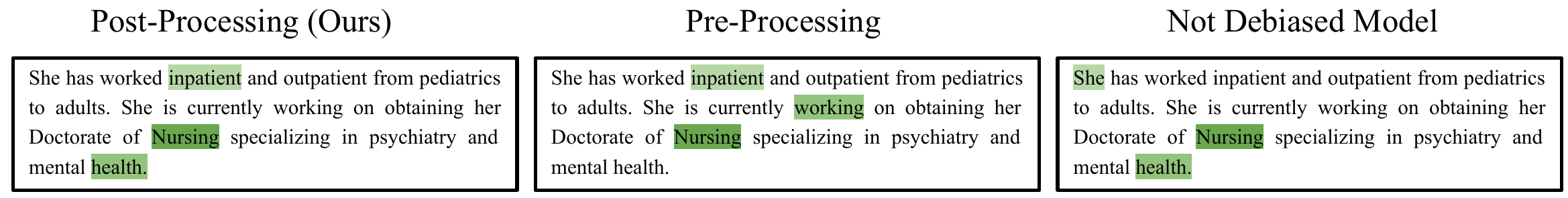}
    \caption{Qualitative results from the non-tabular data experiment on the job classification task based on bio texts. Green regions are the top three words used by the model for its prediction based on the attention weights. While the Not Debiased Model mostly focuses on gendered words, our method focused on profession-based words, such as R.N. (Registered Nurse), to correctly predict ``nurse.''}
    \label{NLP_qualitative}
\end{figure*}
In order to measure the bias, we used the same measure as in~\cite{de2019bias} which is based on the equality of opportunity notion of fairness~\cite{NIPS2016_9d268236} and reported the True Positive Rate Difference (TPRD) for each occupation amongst different genders. As shown in Table~\ref{bios_results}, our post-processing mitigation technique provides lower TRPD while being more accurate, followed by the technique that masks the gendered words before training. Although both methods reduce the bias compared to a model trained on raw bios without applying any mask or invariance to gendered words, our post-processing method is more effective. Fig.~\ref{NLP_qualitative} also highlights qualitative differences between models in terms of their most attentive features for the prediction task. As shown in the results, our post-processing technique is able to use more meaningful words, such as R.N. (registered nurse) to predict the outcome label nurse compared to both baselines, while the non-debiased model focuses on gendered words.

\section{Related Work}
\para{Fairness.} The research in fairness concerns itself with various topics, such as defining fairness metrics, proposing solutions for bias mitigation, and analyzing existing harms in various systems~\cite{10.1145/3457607}. In this work, we utilized different metrics that were introduced previously, such as statistical parity~\cite{Dwork:2012:FTA:2090236.2090255}, equality of opportunity and equalized odds~\cite{NIPS2016_9d268236}, to measure the amount of bias. We also used different bias mitigation strategies to compare against our mitigation strategy, such as FCRL~\cite{gupta2021controllable}, CVIB~\cite{NEURIPS2018_415185ea}, MIFR~\cite{pmlr-v89-song19a}, adversarial forgetting~\cite{advforg}, MaxEnt-ARL~\cite{roy2019mitigating}, and LAFTR~\cite{pmlr-v80-madras18a}.  We also utilized concepts and datasets that were analyzing existing biases in NLP systems, such as~\cite{de2019bias} which studied the existing biases in NLP systems on the occupation classification task on the bios dataset.

\para{Interpretability.} In this work, we introduced an attribution framework based on the attention weights that can analyze fairness and accuracy of the models at the same time and reason about the importance of each feature on fairness and accuracy. There is a body of work in NLP literature that tried to analyze the effect of the attention weights on interpretability of the model~\cite{wiegreffe-pinter-2019-attention,jain-wallace-2019-attention,serrano-smith-2019-attention}. Other work also utilized attention weights to define an attribution score to be able to reason about how transformer models such as BERT work~\cite{hao2021self}. Notice that although \citet{jain-wallace-2019-attention} claim that attention might not be explanation, a body of work has proved otherwise including~\cite{wiegreffe-pinter-2019-attention} in which authors directly target the work in \citet{jain-wallace-2019-attention} and analyze in detail the problems associated with this study. In our work, we also find that attention can be useful and can extract meaningful information which can be beneficial in many aspects. In addition,~\citet{NEURIPS2020_92650b2e} analyze the effect of the attention weights in transformer models for bias analysis in language models. However, their approach is different and has a more causal take on investigating the bias. Their study is specific to language models and does not necessarily apply to broader tasks and existing fairness definitions. Aside from interpretability and fairness, we utilized concepts from the NLP literature for designing our attention-based model that can be applicable to tabular data~\cite{NIPS2017_3f5ee243,zhou-etal-2016-attention}.
 
\section{Discussion}
In this work, we analyzed how attention weights contribute to fairness and accuracy of a predictive model. To do so, we proposed an attribution method that leverages the  attention mechanism and showed the effectiveness of this approach on both tabular and text data. Using this interpretable attribution framework we then introduced a post-processing bias mitigation strategy based on attention weight manipulation. We validated the proposed framework by conducting experiments with different baselines, as well as fairness metrics, and different data modalities.

Although our work can have a positive impact in allowing to reason about fairness and accuracy of models and  reduce their bias, it can also have negative societal consequences if used unethically. For instance, it has been previously shown that interpretability frameworks can be used as a means for fairwashing which is when malicious users generate fake explanations for their unfair decisions to justify them \cite{pmlr-v119-anders20a}. In addition, previously it has been shown that interpratability frameworks are vulnerable against adversarial attacks \cite{slack2020fooling}. We acknowledge that our framework may also be targeted by malicious users for malicious intent that can manipulate attention weights to either generate fake explanations or unfair outcomes. An important future direction can be to analyze and improve robustness of our framework along with others. 


\bibliography{aaai22}

\clearpage
\appendix
\section{Appendix}
We included additional bias mitigation results using other fairness metrics, such as equality of opportunity and equalized odds on both of the Adult and Heritage Health datasets in this supplementary material. We also included additional post-processing results along with additional qualitative results both for the tabular and non-tabular dataset experiments. More details can be found under each sub-section.
\subsection{Results on Tabular Data}
\begin{figure*}[h]
\begin{subfigure}[b]{0.49\textwidth}
\includegraphics[width=\textwidth,trim=0cm 0cm 1.8cm 1.0cm,clip=true]{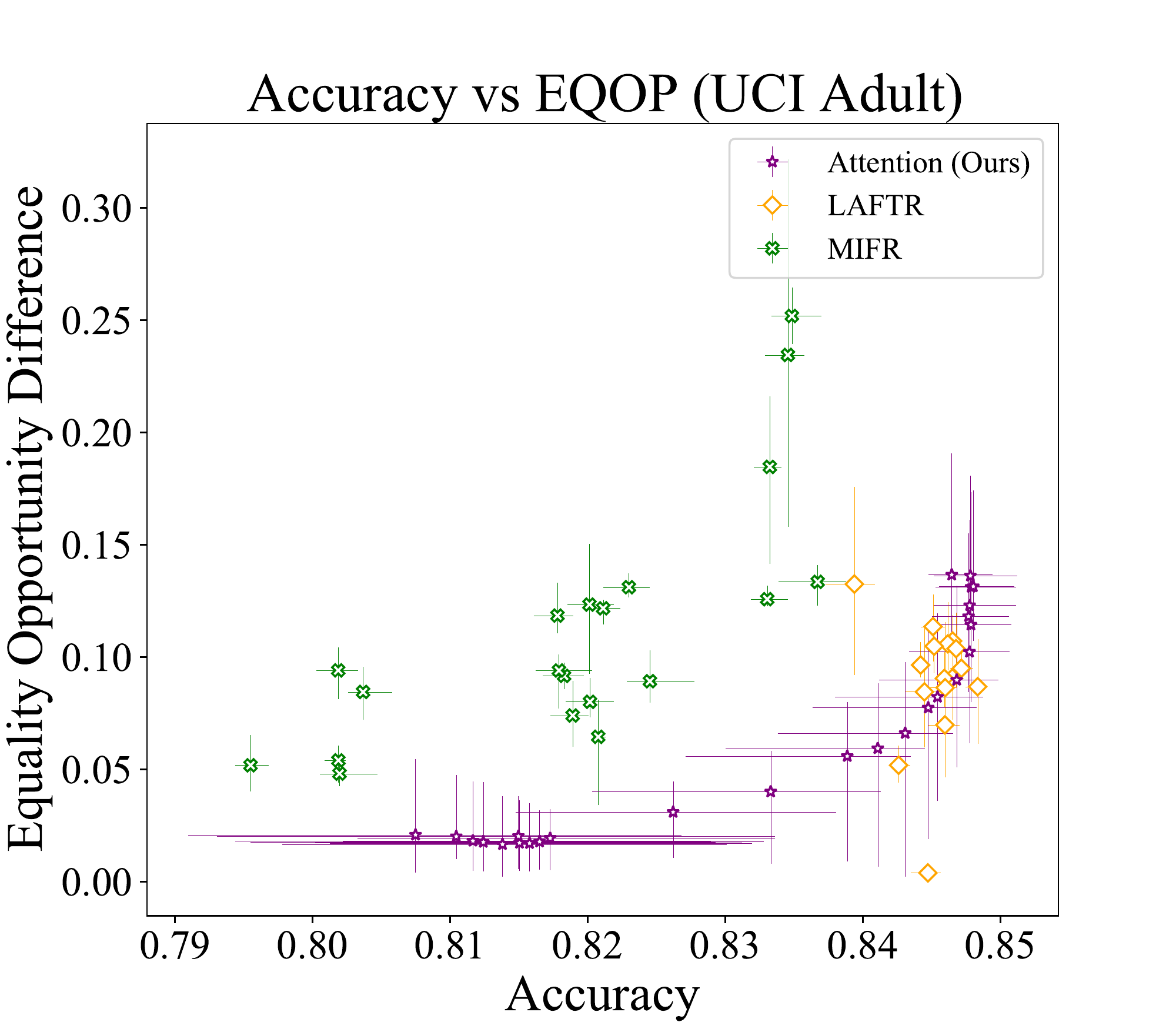}
\end{subfigure}
\begin{subfigure}[b]{0.49\textwidth}
\includegraphics[width=\textwidth,trim=0cm 0cm 1.8cm 1cm,clip=true]{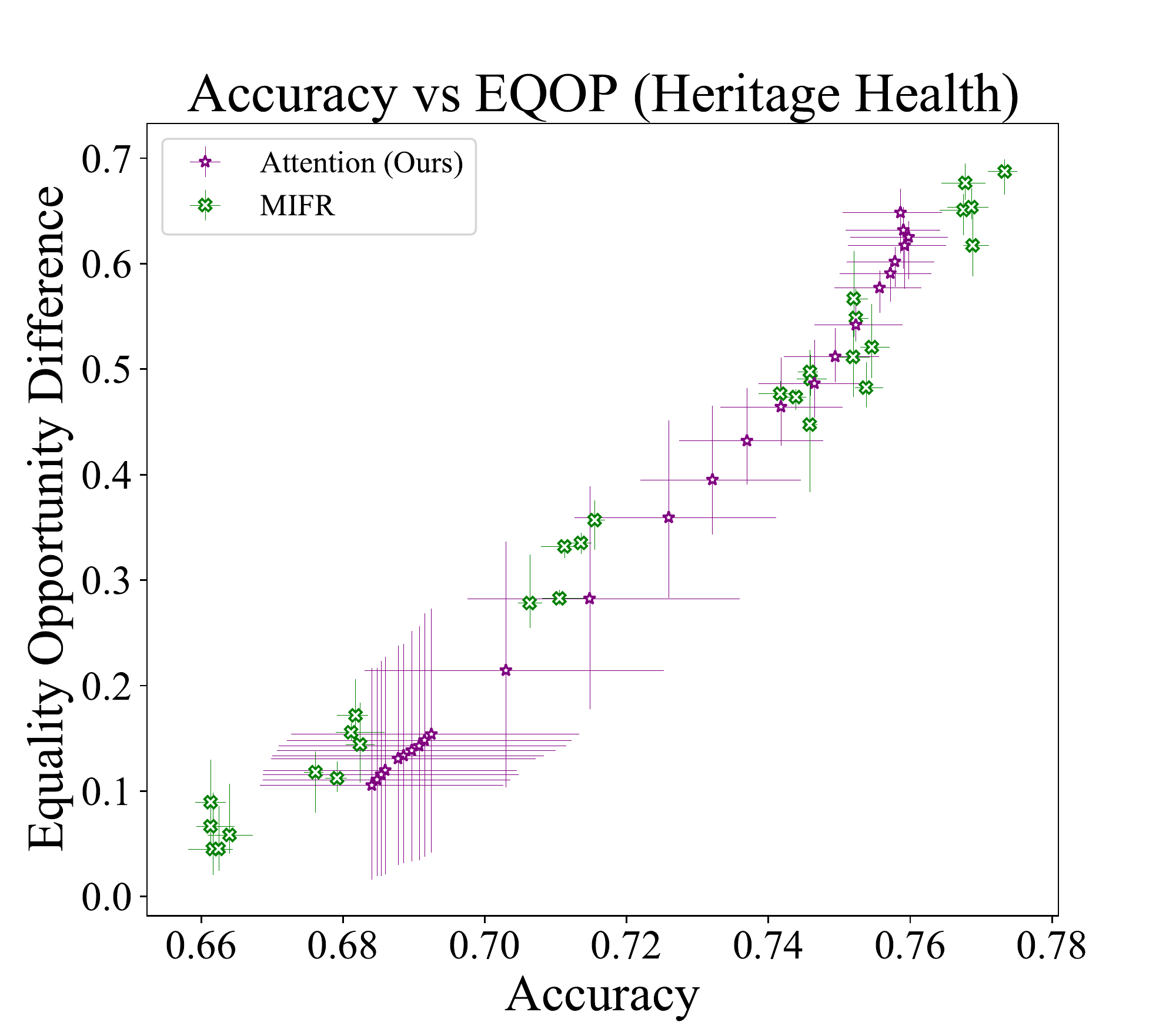}
\end{subfigure}
\caption{Accuracy vs equality of opportunity curves for UCI Adult and Heritage Health datasets.}
\label{mitig_results_eqop}
\end{figure*}

\begin{figure*}[h]
\begin{subfigure}[b]{0.49\textwidth}
\includegraphics[width=\textwidth,trim=0cm 0cm 1.8cm 1.0cm,clip=true]{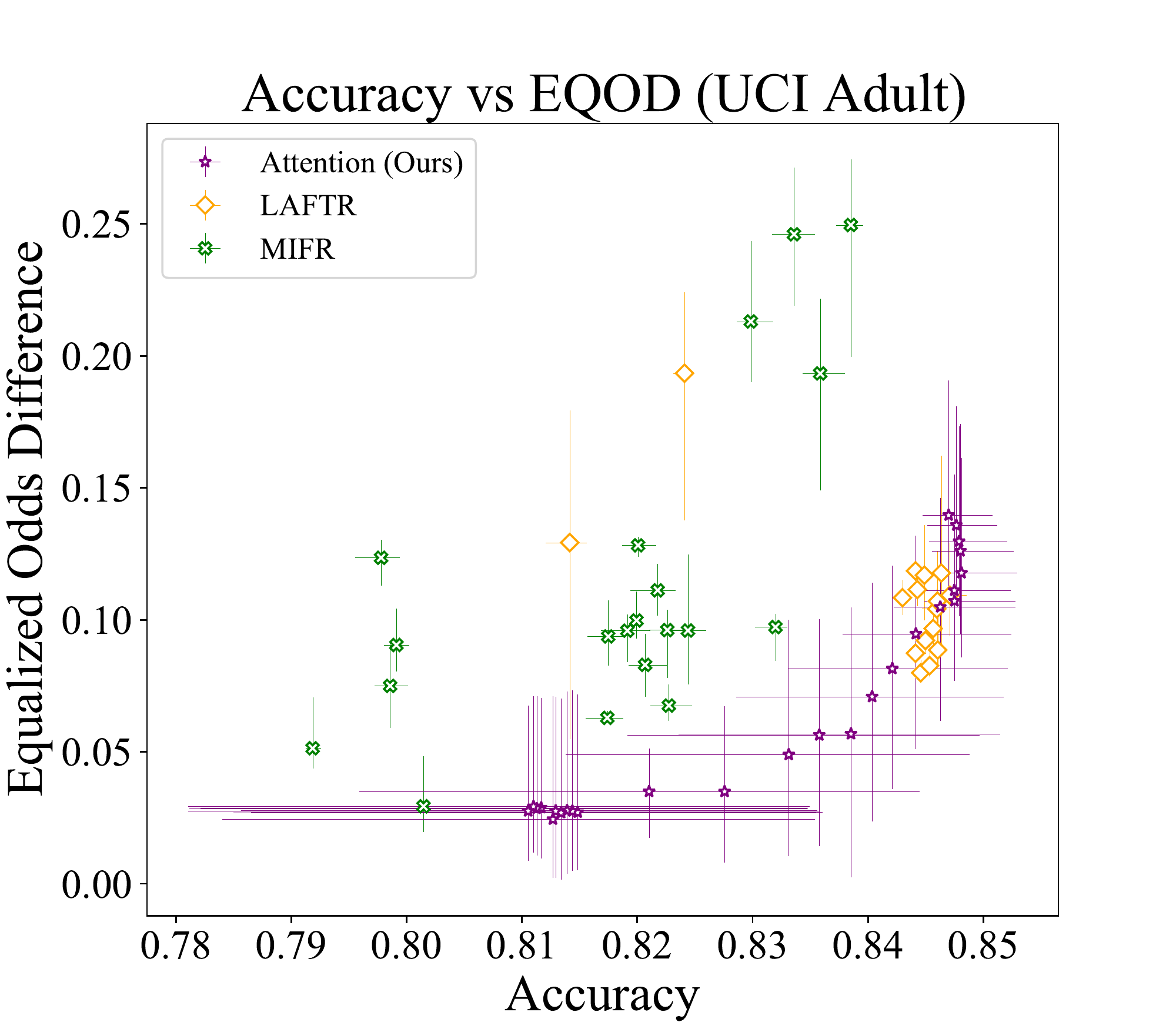}
\end{subfigure}
\begin{subfigure}[b]{0.49\textwidth}
\includegraphics[width=\textwidth,trim=0cm 0cm 1.8cm 1cm,clip=true]{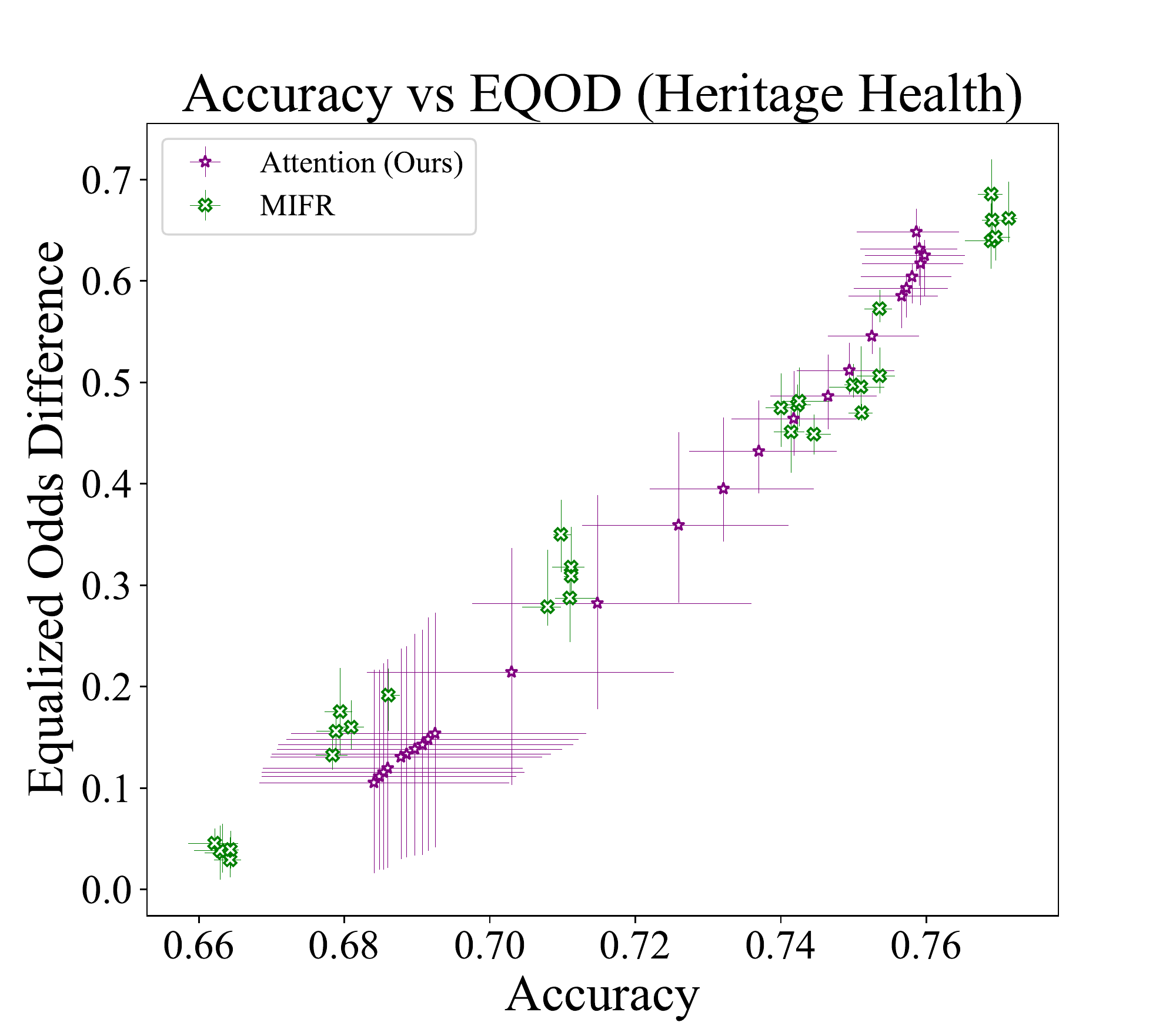}
\end{subfigure}
\caption{Accuracy vs equalized odds curves for UCI Adult and Heritage Health datasets.}
\label{mitig_results_eqod}
\end{figure*}

Here, we show the results of our mitigation framework considering equality of opportunity and equalized odds notions of fairness. We included baselines that were applicable for these notions. Notice not all the baselines we used in our previous analysis for statistical parity were applicable for equality of opportunity and equalized odds notions of fairness; thus, we only included the applicable ones. In addition, LAFTR is only applicable when the sensitive attribute is a binary variable, so it was not applicable to be included in the analysis for the heritage health data where the sensitive attribute is non-binary. Results of these analysis is shown in Figures~\ref{mitig_results_eqop} and \ref{mitig_results_eqod}. We once again show competitive and comparable results to other baseline methods, while having the advantage of being interpretable and not requiring multiple trainings to satisfy different fairness notions or fairness on different sensitive attributes. Our framework is also flexible for different fairness measures and can be applied to binary or non-binary sensitive features.

\begin{figure*}[h]
\centering
\begin{subfigure}[b]{0.44\textwidth}
\includegraphics[width=\textwidth,trim=7cm 2cm 5cm 4cm,clip=true]{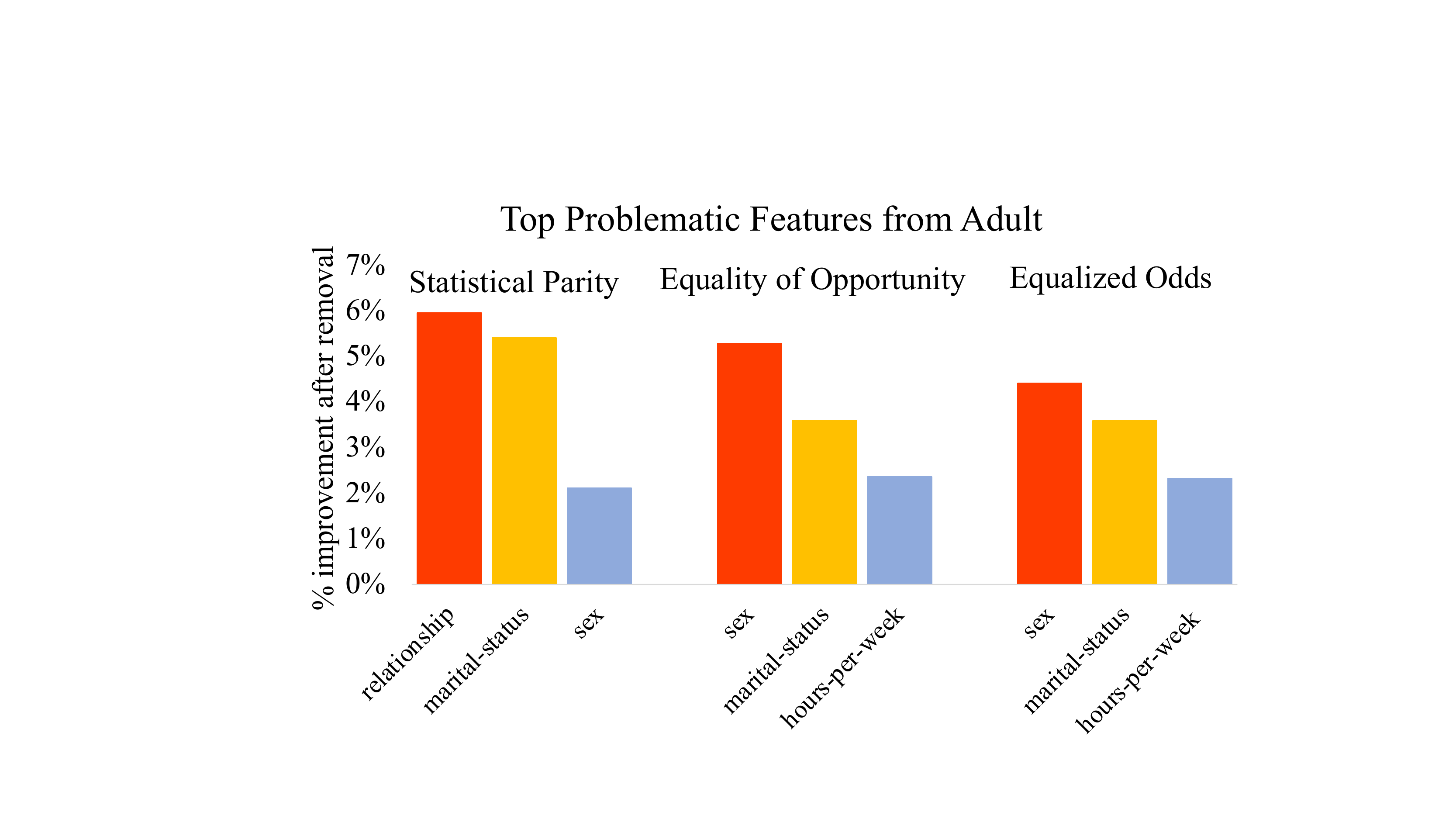}
\end{subfigure}
\begin{subfigure}[b]{0.50\textwidth}
\includegraphics[width=\textwidth,trim=4cm 3cm 4cm 3cm,clip=true]{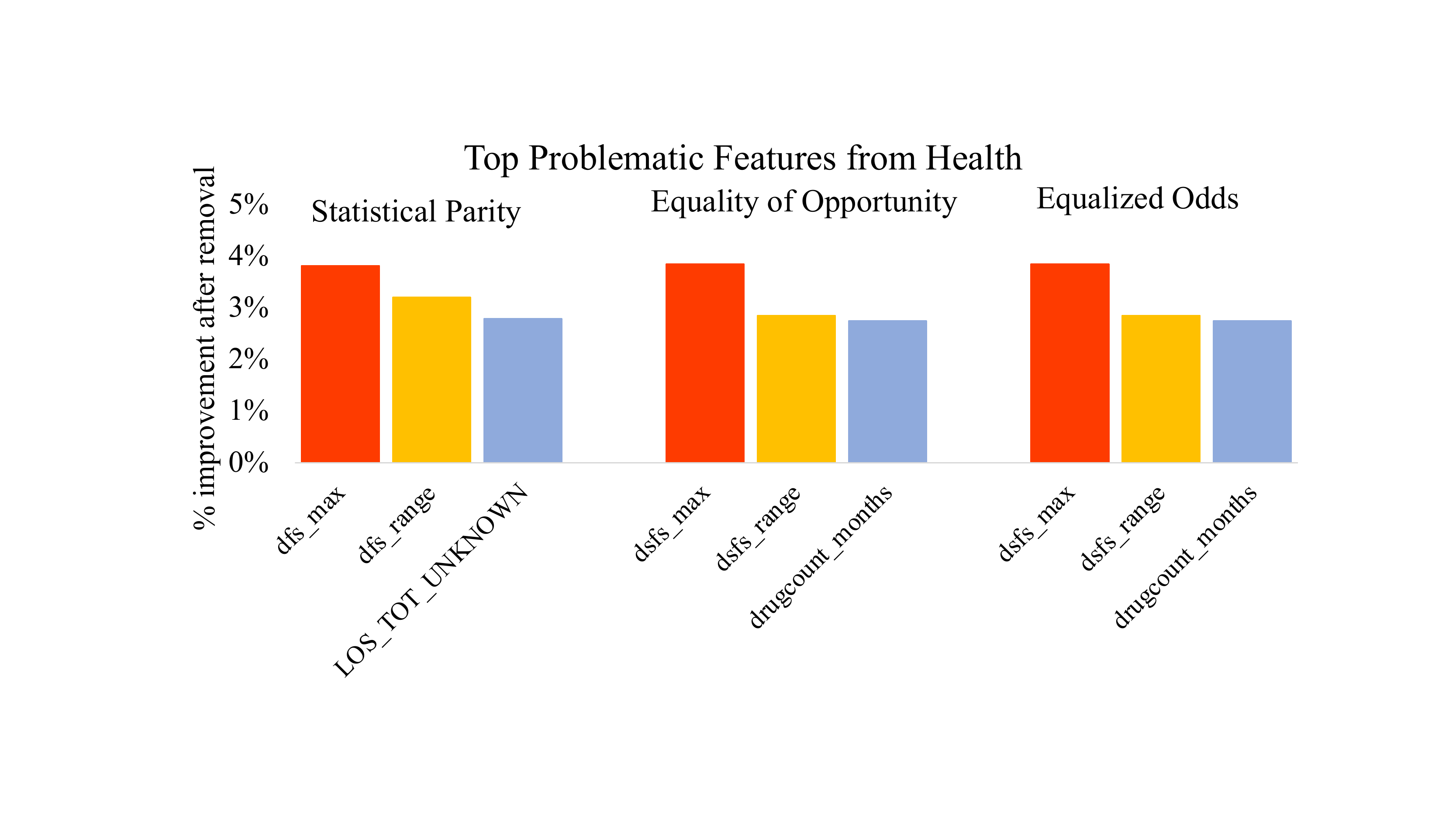}
\end{subfigure}
\caption{Top three features for each fairness definition removing which caused the most benefit in improving the corresponding fairness definition. The percentage of improvement upon removal is marked on the $y$-axis for adult and heritage health datasets.}
\label{features}
\end{figure*}

In addition, we show how different features contribute differently under different fairness notions. Fig.~\ref{features} demonstrates the top three features that contribute to unfairness the most along with the percentages of the fairness improvement upon their removal for each of the fairness notions. As observed from the results, while equality of opportunity and equalized odds are similar in terms of their problematic features, statistical parity has different trends. This is also expected as equality of opportunity and equalized odds are similar fairness notions in nature compared to statistical parity.

We also compared our mitigation strategy with the Hardt etl al. post-processing approach~\cite{NIPS2016_9d268236}. Using this post-processing implementation~\footnote{\url{https://fairlearn.org}}, we obtained the optimal solution that tries to satisfy different fairness notions subject to accuracy constraints. For our results, we put the results from zeroing out all the attention weights corresponding to the problematic features that were detected from our interpretability framework. However, notice that since our mitigation strategy can control different trade-offs we can have different results depending on the scenario. Here, we reported the results from zeroing out the problematic attention weights that is targeting fairness mostly. From the results demonstrated in Tables~\ref{adult_post_proc} and~\ref{health_post_proc}, we can see comparable numbers to those obtained from~\cite{NIPS2016_9d268236}. This again shows that our interpretability framework yet again captures the correct responsible features and that the mitigation strategy works as expected.

\subsection{Results on non-tabular Data}
\begin{figure*}[h]
    \centering
      \includegraphics[width=\textwidth,trim=0cm 6cm 0cm 1cm,clip=true]{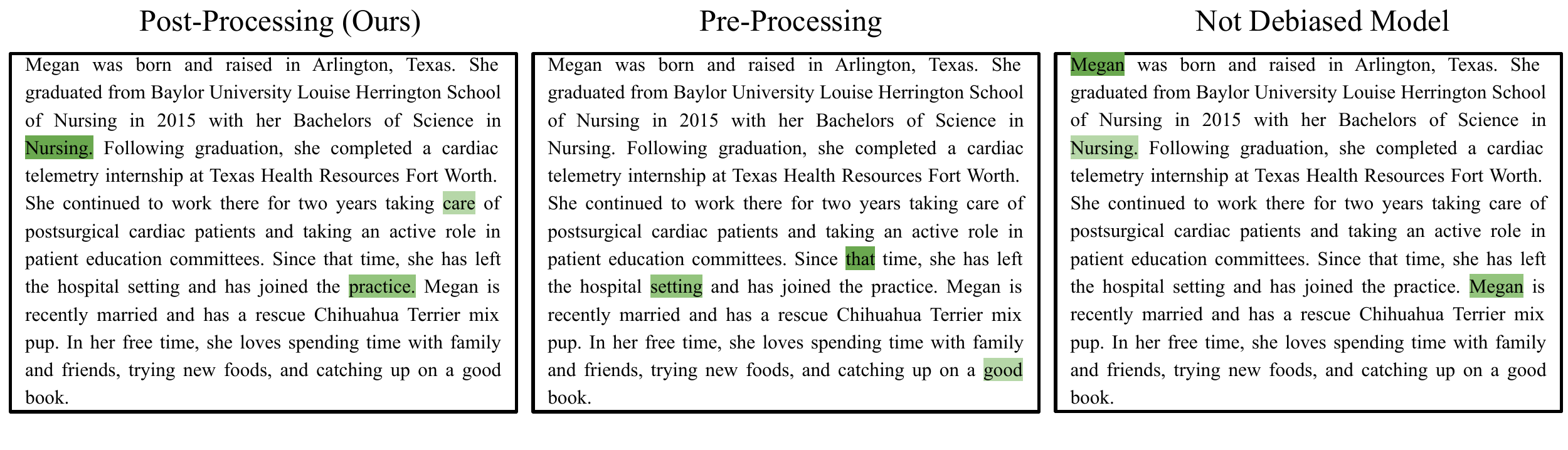}
    \caption{Additional qualitative results from the non-tabular data experiment on the job classification task based on the bio texts. Green regions represent top three words that the model used for its prediction based on the attention weights.}
    \label{fig:NLP_qualitative_appendix}
\end{figure*}

We also included some additional qualitative results from the experiments on non-tabular data in Fig.~\ref{fig:NLP_qualitative_appendix}.

\subsection{Interpreting Fairness with Attention}
\begin{figure*}[h]
\centering
\begin{subfigure}[b]{0.45\textwidth}
\includegraphics[width=\textwidth,trim=2.5cm 0cm 4.5cm 1.0cm,clip=true]{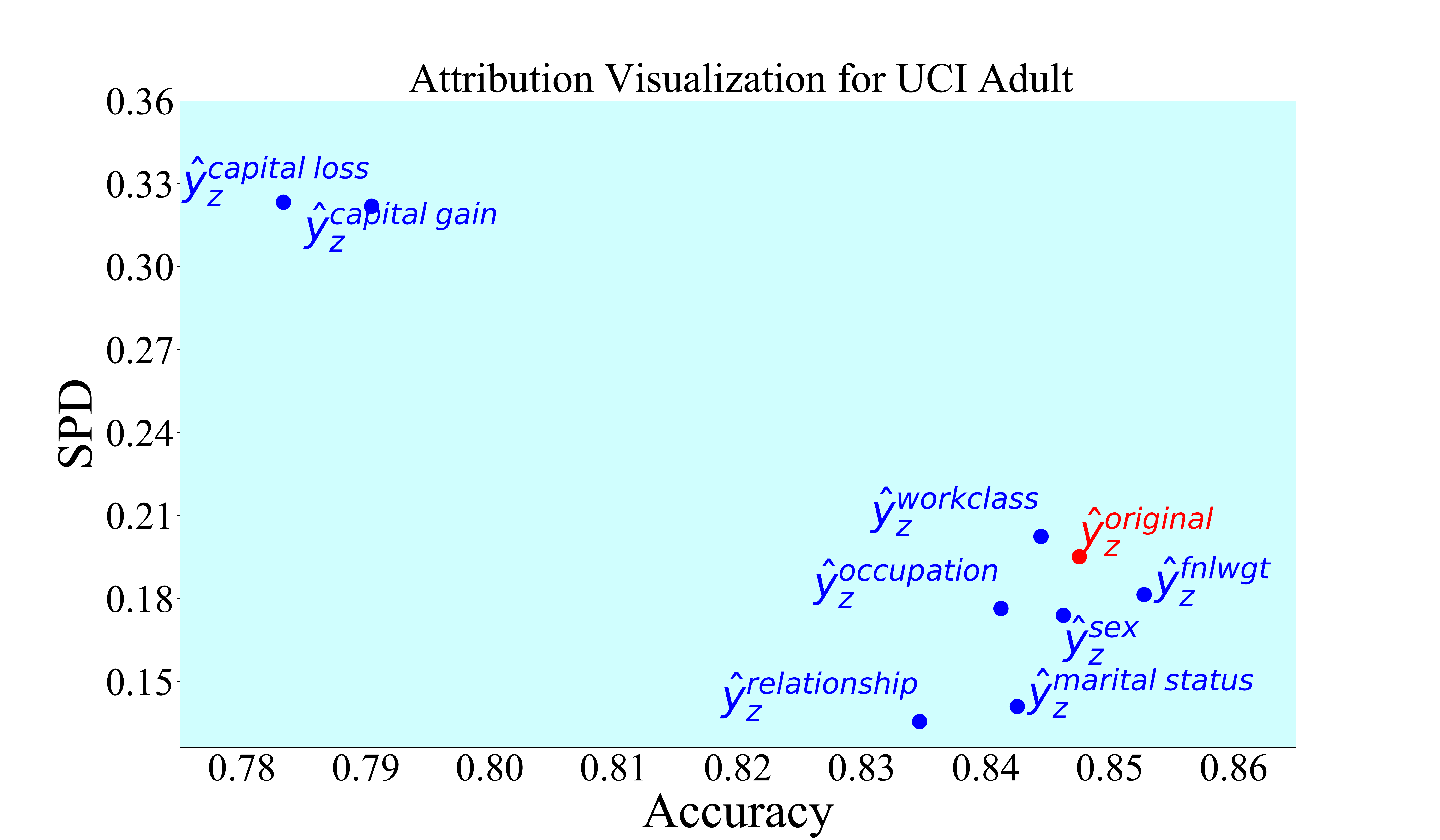}
\end{subfigure}
\begin{subfigure}[b]{0.45\textwidth}
\includegraphics[width=\textwidth,trim=2.4cm 0cm 4.5cm 1.0cm,clip=true]{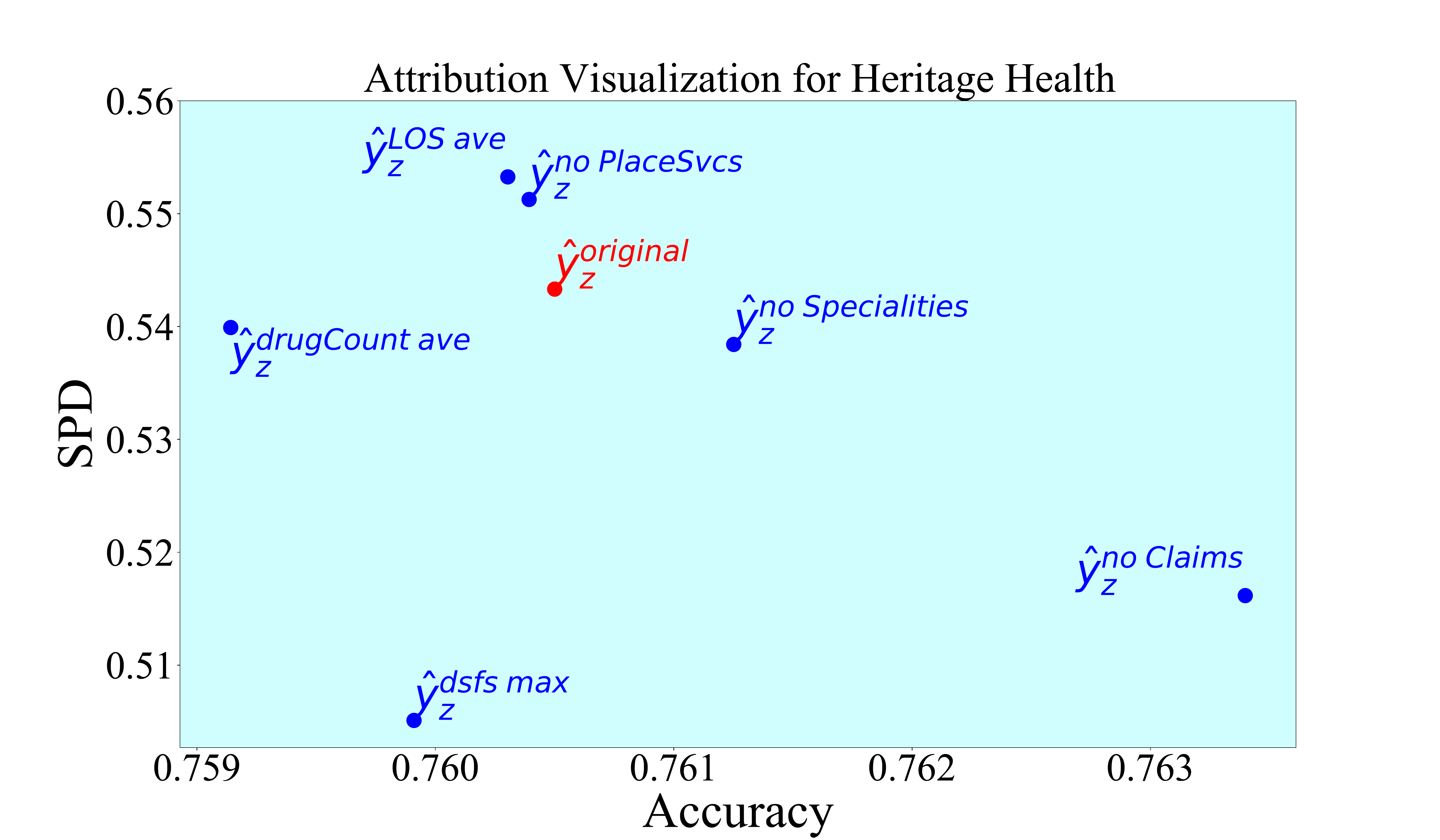}
\end{subfigure}
\caption{Results from the real-world datasets. Note that in our $\hat{y}_z$ notation we replaced indexes with actual feature names for clarity in these results on real-world datasets as there is not one universal indexing schema, but the feature names are more universal and discriptive for this case. Labels on the points represent the feature name that was removed (zeroed out) according to our $\hat{y}_z$ notation. The results show how the accuracy and fairness of the model (in terms of statistical parity difference) change by exclusion of each feature.}
\label{fig:interpret_results_2}
\end{figure*}

Fig.~\ref{fig:interpret_results_2} shows results on a subset of the features from the \textit{UCI Adult} and \textit{Heritage Health} datasets (to keep the plots uncluttered and readable, we incorporated the most interesting features in the plot), and  provide some intuition about how different features in these datasets contribute to the model fairness and accuracy. While features such as {\em capital gain} and {\em capital loss} in the \textit{UCI Adult} dataset are responsible for improving accuracy and reducing bias, we can observe  that features such as {\em relationship} or {\em marital status}, which can be indirectly correlated with the feature {\em sex}, have a negative impact on fairness. For the \textit{Heritage Health} dataset, including the features {\em drugCount ave} and {\em dsfs max} provide accuracy gains but at the expense of fairness, while including {\em no Claims} and {\em no Specialities} negatively impact both accuracy and fairness.

\subsection{Information on Datasets and Features}
More details about each of the datasets along with the descriptions of each feature for the Adult dataset can be found at\footnote{https://archive.ics.uci.edu/ml/datasets/adult} and for the Heritage Health dataset can be found at
\footnote{https://www.kaggle.com/c/hhp}. In our qualitative results, we used the feature names as marked in these datasets. If the names or acronyms are unclear kindly reference to the references mentioned for more detailed description for each of the features. Although most of the features in the Adult datasets are self-descriptive, Heritage Health dataset includes some abbreviations that we list in Table~\ref{feature_abbriv} for the ease of interpreting each feature's meaning.

\begin{table}[h]
\centering
\begin{tabular}{|c|c|}
\hline
\textbf{Abbreviation}                     & \textbf{Meaning}                                                         \\ \hline
\multirow{1}{*}{PlaceSvcs}      & Place where the member was treated.         
                                                          \\ \hline
\multirow{1}{*}{LOS}   & Length of stay.                  
                              \\ \hline
\multirow{1}{*}{dsfs} & Days since first service that year.            
                                           \\ \hline
\end{tabular}
\caption{Some abbreviations used in Heritage Health dataset's feature names. These abbreviations are listed for clarity of interpreting each feature's meaning specifically in our qualitative analysis or attribution visualizations.}
\label{feature_abbriv}
\end{table}

\begin{table*}[h]
\centering
\scalebox{0.9}{
    \begin{tabular}{c | c c  | c c |cc}
        \toprule
        & Accuracy & SPD   & Accuracy & EQOP & Accuracy & EQOD\\
        \midrule
        Attention (Ours)&\textbf{0.77 (0.006)}&\textbf{0.012 (0.003)}&0.81 (0.013)&\textbf{0.020 (0.019)}&\textbf{0.81 (0.021)}&\textbf{0.027 (0.023)}\\
        \midrule
        Hardt et al.&\textbf{0.77 (0.012)}&0.013 (0.005)&\textbf{0.83 (0.005)}&0.064 (0.016)&\textbf{0.81 (0.007)}&0.047 (0.014)\\
        \bottomrule
    \end{tabular}}
    \caption{Adult results on post-processing approach from Hardt et al. vs our attention method when all problematic features are zeroed out.}
    \label{adult_post_proc}
    \vspace{-1em}
\end{table*}

\begin{table*}[h]
\centering
\scalebox{0.9}{
    \begin{tabular}{c | c c  | c c |cc}
        \toprule
        & Accuracy & SPD  & Accuracy & EQOP & Accuracy & EQOD\\
        \midrule
        Attention (Ours)&\textbf{0.68 (0.004)}&\textbf{0.04 (0.015)}&0.68 (0.015)&\textbf{0.15 (0.085)}&0.68 (0.015)&\textbf{0.10 (0.085)}\\
        \midrule
        Hardt et al.&\textbf{0.68 (0.005)}&0.05 (0.018)&\textbf{0.75 (0.001)}&0.20 (0.033)&\textbf{0.69 (0.012)}&0.19 (0.031)\\
        \bottomrule
    \end{tabular}
    }
    \caption{Heritage Health results on post-processing approach from Hardt et al. vs our attention method when all problematic features are zeroed out.}
    \label{health_post_proc}
    \vspace{-1em}
\end{table*}

\end{document}